\def\year{2021}\relax
\documentclass[letterpaper]{article} % DO NOT CHANGE THIS
\usepackage{aaai21}  % DO NOT CHANGE THIS
\usepackage{times}  % DO NOT CHANGE THIS
\usepackage{helvet} % DO NOT CHANGE THIS
\usepackage{courier}  % DO NOT CHANGE THIS
\usepackage[hyphens]{url}  % DO NOT CHANGE THIS
\usepackage{graphicx} % DO NOT CHANGE THIS
\usepackage{natbib}  % DO NOT CHANGE THIS AND DO NOT ADD ANY OPTIONS TO IT
\usepackage{caption} % DO NOT CHANGE THIS AND DO NOT ADD ANY OPTIONS TO IT
\usepackage{multirow}
\usepackage{amssymb}
\usepackage{bm}
\usepackage{subfigure}
\usepackage{algorithm}
\usepackage{amsmath}
\usepackage{algpseudocode}

\title{Relation-Aware Neighborhood Matching Model for Entity Alignment}
\author {
    Yao Zhu$^1$, 
    Hongzhi Liu$^2$\thanks{Corresponding author},
    Zhonghai Wu$^{3,4}$,
    Yingpeng Du$^2$\\
}
\affiliations {
    $^1$ Center for Data Science, Peking University, Beijing, China \\
    $^2$ School of Software and Microelectronics, Peking University, Beijing, China \\
    $^3$ National Engineering Center of Software Engineering, Peking University, Beijing, China \\
    $^4$ Key Lab of High Confidence Software Technologies (MOE), Peking University, Beijing, China \\
    \{yao.zhu, liuhz, wuzh, dyp1993\}@pku.edu.cn
}

\begin{document}

\maketitle

\begin{abstract}
Entity alignment which aims at linking entities with the same
meaning from different knowledge graphs (KGs) is a vital step for
knowledge fusion. Existing research focused on learning embeddings
of entities by utilizing structural information of KGs for entity
alignment. These methods can aggregate information from 
neighboring nodes but may also bring noise from neighbors.
Most recently, several researchers attempted to compare neighboring
nodes in pairs to enhance the entity alignment. However, they
ignored the relations between entities which are also important for
neighborhood matching. In addition, existing methods paid less
attention to the positive interactions between the entity alignment
and the relation alignment. To deal with these issues, we propose a
novel \underline{R}elation-aware \underline{N}eighborhood
\underline{M}atching model named RNM for entity alignment.
Specifically, we propose to utilize the neighborhood matching to
enhance the entity alignment. Besides comparing neighbor nodes when
matching neighborhood, we also try to explore useful information from the
connected relations. Moreover, an iterative framework is designed to
leverage the positive interactions between the entity alignment and
the relation alignment in a semi-supervised manner. Experimental
results on three real-world datasets demonstrate that the proposed
model RNM performs better than state-of-the-art methods.
\end{abstract}

%, which can be used to enhance the reliability of both the two kinds
%of alignments

\section{Introduction}

In knowledge graphs (KGs), facts are presented as triples of
$(h,r,t)$, indicating there is a relation $r$ from the head entity
$h$ to the tail entity $t$. Real-world KGs such as DBpedia
\cite{lehmann2015dbpedia}, YAGO \cite{suchanek2007yago}, and
Freebase \cite{bollacker2008freebase}, which store a great deal of
knowledge, have been employed in various applications like
recommendation systems \cite{cao2019unifying}, question answering
\cite{huang2019knowledge}, and search engines
\cite{xiong2017explicit}.

However, each individual KG may be incomplete. Since different KGs
are constructed independently from different data sources, they are
usually complementary to each other. Therefore, integrating
heterogeneous knowledge from various KGs has become an urgent issue.
Entity alignment is a vital step for knowledge fusion from different
KGs, which aims at linking entities with the equivalent meaning from
different KGs. The facts can consequently be fused based on the
aligned entities.

Regarding the entity alignment task, most of the existing research
focused on constructing embedding-based models. These methods tried
to embed the entities of KGs into a latent space and calculated the
distances between entity vectors as the evidences for alignment.
TransE \cite{bordes2013translating}, as an effective KG embedding
model, has been widely adopted for entity alignment
\cite{hao2016joint, chen2017multilingual, hao2017Iterative,
sun2018bootstrapping}. To better utilize the information from
neighbors, graph convolutional networks (GCNs) \cite{kipf2017semi}
were utilized to improve the representation learning of entities
\cite{wang2018cross, wu2019relation, ye2019vectorized,
sun2020knowledge}. However, these methods concentrated on learning
comprehensive embeddings for entities, meanwhile, may bring
additional noise from neighbors.

\begin{figure}[t]
    \centering
    \includegraphics[width=0.8\linewidth]{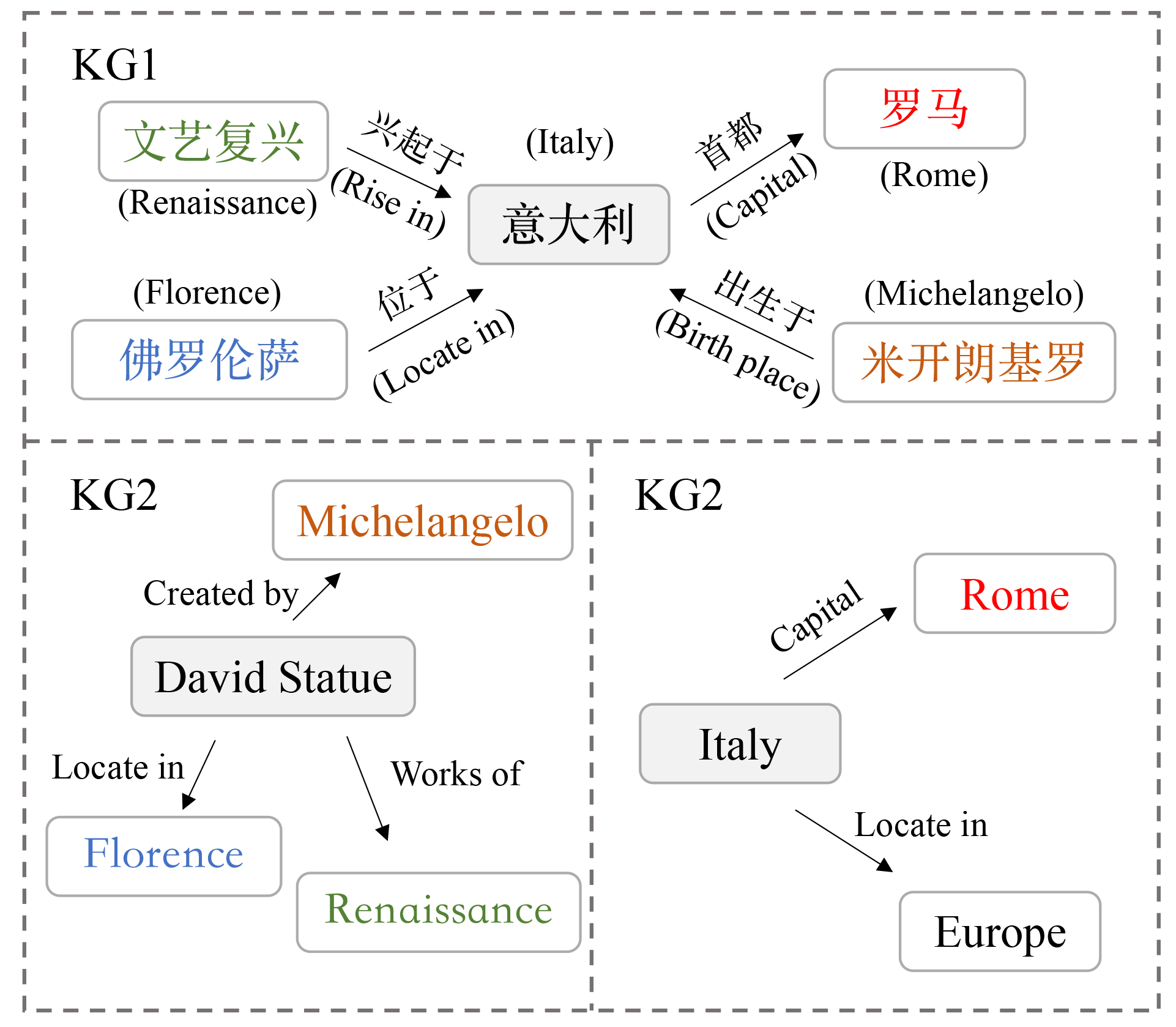}
    \caption{An example to illustrate the importance of relations
    when matching neighborhood. The upper part of the figure is a
    subgraph of KG1 (in Chinese), while the lower parts are two
    subgraphs of KG2 (in English). Assume the entities $Rome$, $Renaissance$, $Florence$ and $Michelangelo$ in the two KGs have been aligned. The entity $Italy$ in KG1 is more likely to be
    misaligned with the entity $David\ Statue$ in KG2 with consideration of the neighbor
    entities while ignoring the connected relations. However, the two $Italy$ entities in different KGs can
    be correctly aligned by matching both the
    relation (Capital) and the neighbor (Rome).}
    \label{example}
\end{figure}

Recently, several studies tried to conduct subgraph matching when
comparing the candidate entity pairs to enhance the alignment
\cite{xu2019cross, wu2020neighborhood}. However, these methods only
compared the neighboring entities but ignored the connected
relations which also contain important information for neighborhood
matching and entity alignment. Moreover, existing methods paid less
attention to the positive interactions between the entity alignment
task and the relation alignment task. Our insights are described as
follows. First, neighborhood matching with relations can enhance the
reliability of entity alignment. Figure \ref{example} shows an
example of the entity alignment with neighborhood matching. Assume
the entities $Rome$, $Renaissance$, $Florence$ and $Michelangelo$
in the two KGs have been aligned. If we only
consider the neighboring entities when matching subgraphs, the
entity $Italy$ (in Chinese) in KG1 is more likely to be misaligned
with the entity $David\ Statue$ in KG2. However, if we compare the
connected relations at the same time and consider the 1-to-1
property of relation $capital$, the entity $Italy$ can be correctly
aligned crossing two KGs. This implies that relations play a
significant role in neighborhood matching not only for the semantic
meaning but also for the mapping property. Second, relation
alignments can help to find the alignments of entities, and on the
other hand, entity alignments can also assist the relation alignment
task. Specifically, the entity alignment can be inferred based on
the neighboring entities and the linking relations, while the
relation alignment can be inferred from the connected head and tail
entities. Thus, it is reasonable to implement both entity alignment
and relation alignment in a unified framework.

Therefore, in this paper, we propose a novel
\underline{R}elation-aware \underline{N}eighborhood
\underline{M}atching model named RNM for entity alignment. Besides
comparing neighboring entities when matching subgraphs, we also
exploit the semantic information and mapping properties from linking
relations for entity alignment. The semantic information of
relations helps us with the relation matching in neighborhood, while
mapping properties of relations provide the probability of
alignment. Moreover, we design an iterative framework to unify the
entity alignment and the relation alignment, in which the two tasks
can reinforce each other in a semi-supervised manner. Experimental
results on three real-world datasets show that RNM significantly
outperforms several state-of-the-art methods.

The remainder of this paper is organized as follows. First, we
discuss the related work and introduce the problem definition in the
following two sections. Then, we describe the proposed model in
detail. After that, experimental settings and empirical evaluation
results are presented. Finally, we conclude the paper and point out some future work.

\begin{figure*}[]
    \centering
    \includegraphics[width=1.0\linewidth]{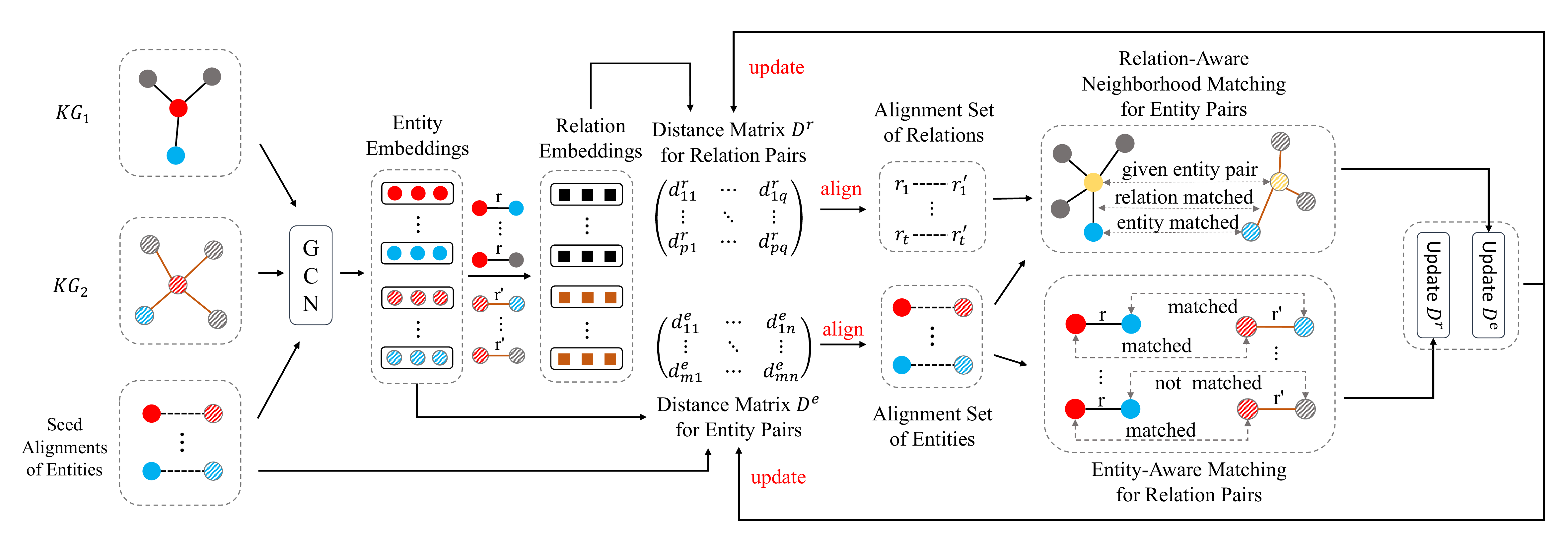}
    \caption{Overall architecture of the proposed model RNM. Assume that there are $m$
    entities in $KG_1$, $n$ entities in $KG_2$, $p$ relations in
    $KG_1$, and $q$ relations in $KG_2$.}
    \label{arc}
\end{figure*}

\section{Related Work}

Most of the existing entity alignment methods focused on embedding
entities from different KGs into the same latent space and measured
the alignment by calculating the distance between entity embeddings.
TransE \cite{bordes2013translating}, as one of the most practical
models for KG embedding, has been adopted for entity alignment.
MTransE \cite{chen2017multilingual} utilized TransE model
to learn entity embeddings for two KGs separately and designed a space transformation mechanism for the
alignment. Instead of training embeddings separately for different
KGs, IPTransE \cite{hao2017Iterative} employed a path-based TransE
model to train the joint knowledge embeddings and proposed an
iterative strategy to expand seed alignments. After that, for better
iteration, BootEA \cite{sun2018bootstrapping} designed a
bootstrapping alignment model based on translational embedding
learning, and used constraints to reduce the error accumulation when
iterating.

Since graph convolutional networks (GCNs) \cite{kipf2017semi} have
achieved remarkable progress in graph learning, some work tried to
apply GCNs to entity alignment for better representation learning.
\citeauthor{wang2018cross} (\citeyear{wang2018cross}) proposed a
GCN-Align model for entity alignment which trained GCNs to embed
entities of each KG into a unified vector space. After that, relations
were taken into account for entity alignment. HGCN \cite{wu2019jointly} 
jointly learned both entity and relation representations 
via a GCN-based framework and RDGCN \cite{wu2019relation} constructed 
a dual relation graph for embedding learning. 
Moreover, AliNet \cite{sun2020knowledge} improved GCNs by aggregating
multi-hop neighborhood with gated strategy and attention
mechanism. These methods tried to make use of the structural and
neighborhood information to learn better representations of
entities. However, they may also bring in some noise from neighbors,
which could degrade the performance of alignment.

More recently, some researchers attempted to employ subgraph
matching for better entity alignments. \citeauthor{xu2019cross}
(\citeyear{xu2019cross}) formulated the KG-alignment task as a graph
matching problem by introducing a local sub-graph for each entity.
NMN \cite{wu2020neighborhood} was a cross-graph neighborhood matching
model which jointly encoded the difference of neighborhood for
entity pairs. However, these methods only took the neighboring
entities for comparison but ignored the connected relations which are
also important for subgraph matching. Thus, in this paper, we
propose a novel relation-aware neighborhood matching model which
explores the semantic information and mapping properties of
relations when conducting subgraph matching. Moreover, entities and
relations are iteratively aligned in our model to make these two
tasks reinforce each other.

\section{Problem Definition}

The entity alignment and the relation alignment are two related
tasks for knowledge fusion.

Formally, a KG can be denoted as $G = (E,R,T)$, where $E$, $R$, and
$T$ are the sets of entities, relations, and triples, respectively.
Given two heterogeneous KGs to be fused, which are $G_1 = (E_1, R_1,
T_1)$ and $G_2 = (E_2, R_2, T_2)$, we assume there is a set of
pre-aligned entity pairs between the two KGs, which is defined as
$\mathbb{L} = \{(e_1,e_2) | e_1 \in E_1, e_2 \in E_2, e_1\ equals\
to\ e_2\}$.

For the entity alignment task, our goal is to find out the remaining
equivalent entity pairs. For the relation alignment task, our goal
is to find out the relation pairs with the same meaning from the two
given KGs. Note that the relation alignment is an unsupervised task
in this paper.

\section{Proposed Model}

In this section, we will first give an overview of the proposed
model RNM. After that, components of the model will be described in
detail, which are embedding learning for entities and relations,
relation-aware neighborhood matching for entity pairs, and
entity-aware matching for relation pairs. Finally, we will present
the iterative strategy and some implementation details of RNM.

\subsection{Overview of RNM}

Figure \ref{arc} illustrates the overall architecture of the
proposed model RNM. First, given two KGs and a set of seed
alignments of entities, we jointly learn the embeddings of entities
and relations using GCNs with a TransE-like regularizer. After that,
we iteratively align the entities and relations in a semi-supervised
manner. In each iteration, we utilize the graph structure
information to determine new matching pairs of entities and
relations by a relation-aware neighborhood matching module and a
entity-aware entity matching module, respectively.

%we first calculate an entity distance matrix and an relation
%distance matrix based on the embeddings and the aligned pairs of
%entities and relations. Then,

%we first form entity alignment set and relation alignment set based
%on distance matrices computed at the last iteration. Next, we devise
%a relation-aware neighborhood matching module for entity pairs and
%an entity-aware matching strategy for relation pairs. In
%relation-aware neighborhood matching module, besides comparing
%neighboring entities when matching subgraphs, we also exploit the
%semantic information and mapping properties from connected relations
%for alignment. Finally, after several iterations, two updated
%distance matrices will be obtained for entity pairs and relation
%pairs, respectively. Considering both entity alignments and relation
%alignments could make these two tasks reinforce each other in
%matching module by updating the aligned pairs iteratively.

\subsection{Embedding Learning for Entity and Relation}

To align the entities of two KGs, we embed them into the same
latent space to make them comparable. Similarly, we embed the
relations of the two KGs into the same latent space for relation
alignment. To explore the interactions between entities and
relations in the KG, we propose to jointly learn the embeddings of
entities and relations.

%In our model, entity embeddings and relation embeddings will be
%utilized for neighborhood matching. To explore the interactions
%between entities and relations in the KG, we jointly learn the
%embeddings of entities and relations in this part.

\subsubsection{Entity Embedding}
Given two KGs and a set of seed alignments of entities, we utilize
GCNs to embed all the entities of the two KGs into the same latent
space with consideration of the structure information of the two
KGs. Following \cite{xu2019cross, wu2020neighborhood}, we
initialize the entity representations with the pre-trained word
embeddings which can provide useful semantic information of
entities. Moreover, we adopt the highway
strategy \cite{srivastava2015highway} to control the noise in the propagation
procedure of GCNs with multiple layers.

%GCNs are adept at exploiting structural information in the graph and
%have shown the remarkable accomplishments in graph representation
%tasks.

%Embeddings of entities in two KGs will be obtained from the output
%of GCN stated above with multiple layers,

We take the outputs of GCN stated above as the embeddings of
entities, and define the final representations of all entities as
$\tilde{\bm{X}} = \{ \tilde{\bm{x}}_{1}, \tilde{\bm{x}}_{2}, \cdots
\tilde{\bm{x}}_{n} | \tilde{\bm{x}}_{i} \in \mathbb{R}^{\tilde{d}}
\}$, where $\tilde{d}$ denotes the dimension of entity embeddings
and $n$ denotes the number of entities. For an entity pair
$(e_i,e'_j)$ where $e_i \in E_1$ and $e'_j \in E_2$, we define the
distance between them as
\begin{equation}
d(e_i, e'_j) = || \tilde{\bm{x}}_{e_i} - \tilde{\bm{x}}_{e'_j}
||_{1},
\end{equation}
where $|| \cdot ||_1$ denotes the 1-norm measure for vectors.
Smaller $d(e_i,e'_j)$ indicates the higher probability of alignment
between the two entities $e_i$ and $e'_j$.

To embed the entities of two KGs into the same latent space, we take
the seed alignments as training data and design a margin-based loss
function for entity alignment as follows,
\begin{equation}
L_E = \sum_{(p,q)\in \mathbb{L}} \sum_{(p',q')\in \mathbb{L}^{'}}
max\{0, d(p,q) - d(p',q') + \gamma\}, \label{loss_e}
\end{equation}
where $\mathbb{L}$ denotes the set of pre-aligned entity pairs,
$\mathbb{L}'$ is a set of negative alignments upon nearest neighbor
sampling \cite{wu2020neighborhood}, and $\gamma > 0$ denotes the margin. 
The loss function
assumes that the distance between aligned entity pairs should be
close to zero, while the distance between negative samples should be
as far as possible.

\subsubsection{Relation Embedding}
%Besides the entities in KGs, we also try to represent relations in
%KGs as vectors.
In the KG, facts are encoded as triples, i.e., $(h,r,t)$, where $h$
denotes the head entity, $t$ denotes the tail entity, and $r$
denotes the relation from $h$ to $t$. Therefore, the meaning of a
relation is associated with its two connected entities. To leverage
the information of connected entities, we utilize the embeddings of
head entities and tail entities learned from GCNs to represent
relations in the KG, which can be written as follows,
\begin{equation}
\bm{r} =  concat[\bm{g}^h_r, \bm{g}^t_r],
\end{equation}
where $\bm{r} \in \mathbb{R}^{2\tilde{d}}$ denotes the embedding of
the relation $r \in R_1 \cup R_2$, $concat$ means the operation of concatenation, and
$\bm{g}^h_r$ and $\bm{g}^t_r$ denote the average embeddings of all
distinct head entities and tail entities for $r$, respectively.

Moreover, to further explore the translational information for
relations based on triples, inspired by TransE
\cite{bordes2013translating}, we design a regularizer as follows,
\begin{equation}
\Omega_R = \sum_{(h,r,t) \in T_1 \cup T_2} ||
\bm{h}+\bm{W}_{R}\bm{r}-\bm{t} ||_1,
\end{equation}
where $T_1$ and $T_2$ denote the sets of triples for two given KGs
$G_1$ and $G_2$, respectively. $\bm{W}_{R} \in \mathbb{R}^{\tilde{d}
\times 2\tilde{d}}$ denotes the transformation matrix from the
latent relation space to latent entity space, which is the model
parameter to be learned. %, and $|| \cdot ||_1$ is the 1-norm measure.

\subsubsection{Objective function}
To jointly learn the embeddings of entities and relations, we
formulate the objective function as follows,
\begin{equation}
L = L_E + \lambda \cdot \Omega_R, \label{loss_all}
\end{equation}
where $\lambda$ is a trade-off coefficient to balance the loss of
entity alignment and the loss of regularizer with consideration of
the embeddings of relations. Our goal is to minimize the function
above after the pre-training of entity embeddings. In addition, we
utilize Adam \cite{kingma2015adam} for the objective optimization.

\subsection{Relation-Aware Neighborhood Matching}

GCNs aim to aggregate information from neighboring nodes but may
also bring some additional noise from neighbors. To reduce the
impact of these noise, we propose a relation-aware neighborhood
matching model to compare entity pairs. We assume that if two
entities from different KGs have been aligned, then with the
relation of the same meaning, the alignment probability of two
pointing tail entities can be inferred according to the mapping
property of the relation. For instance, 1-to-1 relation can provide
the exact alignment while 1-to-N relation can only show the
probability of $1/N$.

%Thus, at this stage, we propose a novel relation-aware neighborhood
%matching method to compare entity pairs at subgraph aspect.

For each candidate entity pair $(e_i, e'_j)$ where $e_i \in G_1$ and
$e'_j \in G_2$, besides comparing their one-hop neighbor entities in
pairs, we also consider the comparison between connected relations.
Specifically, let $\mathcal{N}_{e_i}$ be the set of one-step
neighbor entities of $e_i$ in $G_1$, and $\mathcal{N}_{e'_j}$ be the
set of one-step neighbor entities of $e'_j$ in $G_2$. For
neighborhood matching with respect to $e_i$ and $e'_j$, we compare
all the entity pairs and the connected relation pairs in $C^e_{ij} =
\{ (n_1,n_2), (r_1,r_2) | n_1 \in \mathcal{N}_{e_i}, n_2 \in
\mathcal{N}_{e'_j}, (e_i, r_1, n_1) \in T_1, (e'_j, r_2, n_2) \in
T_2 \}$, where $T_1$ and $T_2$ are the sets of triples for the two
KGs, respectively. After that, we focus on the matched neighbors
with matched relations which are vital for entity alignment. Thus,
the matched set $M^e_{ij}$ is defined as the subset of $C^e_{ij}$,
in which the elements satisfy $(n_1,n_2) \in \mathbb{L}_e$ and
$(r_1,r_2) \in \mathbb{L}_r$, where $\mathbb{L}_e$ denotes the
alignment set of entities and $\mathbb{L}_r$ denotes the
alignment set of relations.

Moreover, mapping properties of connected relations are also important for entity alignment.
Thus, for each matched case in $M^e_{ij}$, we will compute the alignment probability
based on $r_1$, $r_2$ and $n_1$, $n_2$, which can be written as follows,
\begin{equation}
P(r_1,r_2,n_1,n_2) = P(r_1,n_1) \cdot P(r_2,n_2)
\label{ap}
\end{equation}
where
\begin{equation}
P(r_1,n_1) = \frac{1}{|\{e|(e,r_1,n_1) \in T_1\}|}
\end{equation}
and
\begin{equation}
P(r_2,n_2) = \frac{1}{|\{e|(e,r_2,n_2) \in T_2\}|}.
\end{equation}
$P(r_1,n_1)$ and $P(r_2,n_2)$ denote the mapping probability with respect to
the corresponding relation and neighbor entity, respectively. Thus, we can update the
distance between two entities as follows,
\begin{equation}
d^e_{ij} = || \tilde{\bm{x}}_{e_i} - \tilde{\bm{x}}_{e'_j} ||_{1} -
\lambda_e \cdot \frac{\sum_{M^e_{ij}}
P(r_1,r_2,n_1,n_2)}{|\mathcal{N}_{e_i}| + |\mathcal{N}_{e'_j}|},
\label{eq:de}
\end{equation}
where $\lambda_e$ is a hyper-parameter to control the tradeoff
between the embedding distance and the matching score. Greater
matching score indicates the higher probability of alignment for the
candidate entity pair.

\subsection{Entity-Aware Relation Matching}

For two relations from different KGs, we assume that
the more alignments of head entities and tail entities are
at the same time in their associated triples, the more likely the two
relations are with the same meaning.
For a relation $r$, we define
$S_r = \{ (h,t) | (h,r,t) \in T \}$ as the set of its related entity
pairs, where $T$ denotes the set of triples in the given KG. Thus,
given a candidate relation pair $(r_i, r'_j)$ where $r_i$ from $G_1$
and $r'_j$ from $G_2$, we first form the corresponding entity
pair sets $S_{r_i}$ and $S_{r'_j}$. Then, we compare all entity pairs in $C^r_{ij} =
\{(h_1, h_2), (t_1, t_2) | (h_1,t_1) \in S_{r_i}, (h_2,t_2) \in
S_{r'_j} \}$ and define the matching set $M^r_{ij}$ as the subset of
$C^r_{ij}$ where elements meet the conditions of $(h_1,h_2) \in
\mathbb{L}_e$ and $(t_1,t_2) \in \mathbb{L}_e$. Therefore, the
distance between the relation pair $(r_i, r'_j)$ can be updated as
follows,
\begin{equation}
d^r_{ij} = || \bm{r}_i - \bm{r}_j' ||_{1} - \lambda_r \cdot
\frac{|M^r_{ij}|}{|S_{r_i}| + |S_{r'_j}|}, \label{eq:dr}
\end{equation}
where $\lambda_r$ is a tradeoff coefficient. Similar as the distance
measure for the entity pairs, we consider both the embedding
distance and the matching score for relation pairs.

\subsection{Iterative Strategy and Implementation Details}
To make use of the positive interactions between the entity
alignment task and the relation alignment task, we design a
semi-supervised framework in which the entity alignment and the
relation alignment can enhance each other iteratively. Let $D^e \in
\mathbb{R}^{|E_1| \times |E_2|}$ denote the distance matrix for
entity pairs from $KG_1$ to $KG_2$, and $D^r \in \mathbb{R}^{|R_1|
\times |R_2|}$ denote the distance matrix for relation pairs from
$KG_1$ to $KG_2$. Algorithm \ref{is} presents the iterative
strategy of RNM.

The initialization of $D^e$ is defined as follows with the learned
embeddings of entities,
\begin{equation}
d^e_{ij} = \left\{
\begin{aligned}
&\qquad \  0 & (e_i, e'_j) \in \mathbb{L} \\
& \qquad \infty & (e_i, e'_k) \in \mathbb{L}\ \wedge \  j\neq k \\
& || \tilde{\bm{x}}_{e_i} - \tilde{\bm{x}}_{e'_j} ||_{1}  &
otherwise
\end{aligned}
\right. , \label{inide}
\end{equation}
and the initialization of $D^r$ can be written as follows with the
learned embeddings of relations,
\begin{equation}
d^r_{ij} = || \bm{r}_i - \bm{r}_j' ||_{1}. \label{inidr}
\end{equation}

\begin{algorithm}[H]
\caption{Iterative Strategy of RNM} 
\label{is}
\begin{algorithmic}[1]
\Require Entity embeddings $\{\tilde{\bm{x}}_i\}$, relation
embeddings $\{{\bm{r}}_i\}$, seed alignments of entities
$\mathbb{L}$, maximum number of iterations $T$.

\Ensure $D^e$ (distance matrix of entities) and $D^r$ (distance
matrix of relations).

\State Initialize $D^e$ using Eq. (\ref{inide});

\State Initialize $D^r$ using Eq. (\ref{inidr});

\Repeat

\State Update alignment sets according to
Algorithm \ref{pred};
\State Update $D^e$ using Eq. (\ref{eq:de})
with the consideration of relation-aware neighborhood matching;
\State Update $D^r$ using Eq. (\ref{eq:dr}) with the consideration
of entity-aware matching;

\Until{$D^e$, $D^r$ are converged or the iteration reaches $T$}
\\
\Return $D^e$, $D^r$
\end{algorithmic}
\end{algorithm}

$D^e$ and $D^r$ can be utilized for alignment ranking or alignment set generation. The method for generating or updating the alignment sets is shown in Algorithm \ref{pred}.

\begin{algorithm}[H]
\caption{Update Alignment Sets} 
\label{pred}
\begin{algorithmic}[1]
\Require $D^e$ (distance matrix of entities), $D^r$ (distance matrix
of relations), distance threshold $\delta_e$ and $\delta_r$ \Ensure
$\mathbb{L}_e$ (alignment set of entities), $\mathbb{L}_r$
(alignment set of relations). \State Initialize
$\mathbb{L}_e \leftarrow \emptyset$, $\mathbb{L}_r \leftarrow
\emptyset$; \For {each entity $e_i$ in $KG_1$} \State $j =
{\arg\min}_{j}\ {d^e_{ij}}$ \  // find the nearest entity in $KG_2$
\If {$d^e_{ij} < \delta_e$} \State $\mathbb{L}_e \leftarrow
\mathbb{L}_e \cup (e_i,e'_j)$ \EndIf \EndFor \For {each relation
$r_i$ in $KG_1$} \State $j = {\arg\min}_{j}\ {d^r_{ij}}$ \ // find
the nearest relation in $KG_2$ \If {$d^r_{ij} < \delta_r$} \State
$\mathbb{L}_r \leftarrow \mathbb{L}_r \cup (r_i,r'_j)$ \EndIf
\EndFor \State For the conflicts in $\mathbb{L}_e$ or $\mathbb{L}_r$,
we will choose the pair with smaller distance;
\\
\Return $\mathbb{L}_e$, $\mathbb{L}_r$
\end{algorithmic}
\end{algorithm}

In addition, we introduce the reverse relations to enrich the KGs.
For instance, for the fact (Tokyo, \emph{CapitalOf}, Japan), we will
also build another triple (Japan, \emph{CapitalOf}$^{-1}$, Tokyo).
Thus, the set of relations and the set of triples of a given KG will
be accordingly enlarged.

%\subsection{Complexity Analysis}
%
%Let $m$ denote the number of unaligned entities in $KG_1$, $N_c$
%denote the number of matching candidates, $n$ denote the average
%number of neighbors for each entity. The time complexity for
%updating $D^e$ according to Eq. (\ref{eq:de}) is $O(mnN_c)$ since we
%could just judge whether the counterpart of the neighbor is in the
%neighbor set of the candidate entity instead of comparing in pairs.
%Similarly, Let $m_r$ denote the number of relations in $KG_1$,
%$N^r_c$ denote the number of matching candidates, $n_r$ denote the
%average number of related triples for each relation. The time
%complexity for updating $D^r$ according to Eq. (\ref{eq:dr}) is
%$O(m_r n_r N^r_c)$.

\section{Experiments}

\subsection{Experimental Setup}

\subsubsection{Datasets}

To evaluate the performance of the proposed model, we utilize three
cross-lingual datasets from DBP15K as the experimental data. These
datasets are subsets of the large-scale knowledge graph DBpedia
\cite{lehmann2015dbpedia} and are selected from different language
versions including English, Chinese, Japanese, and French. Each
dataset consists of two KGs of different languages and 15,000
aligned entity pairs. Recently, these three datasets have been
widely employed by researchers for entity alignment
\cite{wu2019relation, sun2020knowledge, wu2020neighborhood}. The
statistic details of the datasets are shown in Table \ref{datasets}.

\begin{table}[htb]
\centering \scriptsize
\begin{tabular}{c|c|c|ccc}
\hline \multicolumn{3}{c|}{\textbf{Datasets}}         &
\textbf{Ent.}    & \textbf{Rel.}     & \textbf{Tri.}  \\ \hline
\multicolumn{2}{c|}{\multirow{2}{*}{$DBP15K_{ZH-EN}$}} & Chinese        & 66,469   & 2,830   & 153,929   \\
\multicolumn{2}{c|}{}                       & English        &
98,125   & 2,317   & 237,674   \\ \hline
\multicolumn{2}{c|}{\multirow{2}{*}{$DBP15K_{JA-EN}$}} & Japanese        & 65,744   & 2,043   & 164,373   \\
\multicolumn{2}{c|}{}                       & English        &
95,680   & 2,096   & 233,319   \\ \hline
\multicolumn{2}{c|}{\multirow{2}{*}{$DBP15K_{FR-EN}$}} & French        & 66,858   & 1,379   & 192,191   \\
\multicolumn{2}{c|}{}                       & English        &
105,889   & 2,209   & 278,590  \\ \hline
\end{tabular}
\caption{Statistics of datasets} \label{datasets}
\end{table}

\subsubsection{Experimental Settings}

We employ a 2-layer GCN to learn the entity embeddings. The
dimension of hidden layer in GCN is set as 300. The learning rate is
set to 0.001. Following \cite{wu2020neighborhood}, we first translate 
non-English entity names into English and then
initialize the entity embeddings with the pre-trained word vectors
from Glove model, and the proportion of seed alignments is set as
30\%. Besides, we set the margin $\gamma$ as 1, threshold $\delta_e$
as 5, threshold $\delta_r$ as 3, $\lambda$ as 0.001, $\lambda_e$ as
10, and $\lambda_r$ as 200. We select the nearest 100 entities and
the nearest 20 relations as candidates for matching. The
number of negative samples for each positive one is set as 125, the
maximum number of iterations $T$ is set as 4. We first optimize Eq.
(\ref{loss_e}) for 50 epochs, and then jointly train the embeddings using 
Eq. (\ref{loss_all}) for 10 epochs.

We utilize TensorFlow to implement the proposed model RNM. The
experiments are conducted on a server with two Intel(R) Xeon(R) CPUs
E5-2660 @ 2.20GHz, an NVIDIA Tesla P100 GPU and 16 GB memory.

\begin{table*}[]
\centering \small
\begin{tabular}{c|ccc|ccc|ccc}
\hline
\multirow{2}{*}{Models}    & \multicolumn{3}{|c|}{ZH-EN} & \multicolumn{3}{|c|}{JA-EN}  & \multicolumn{3}{|c}{FR-EN} \\
\cline{2-10}
 & Hits@1    &  Hits@10    & MRR     & Hits@1    &  Hits@10    & MRR    & Hits@1    &  Hits@10    & MRR  \\
\hline
MTransE  \cite{chen2017multilingual}                  & 30.8        & 61.4    & 0.364     & 27.9        & 57.5        & 0.349     & 24.4    & 55.6    & 0.335    \\
IPTransE      \cite{hao2017Iterative}            & 40.6        & 73.5    & 0.516     & 36.7        & 69.3        & 0.474     & 33.3    & 68.5    & 0.451    \\
BootEA    \cite{sun2018bootstrapping}                 & 62.9        & 84.8    & 0.703     & 62.2        & 85.4        & 0.701     & 65.3    & 87.4    & 0.731    \\
AKE    \cite{lin2019guiding}                       & 32.5        & 70.3    & 0.449     & 25.9        & 66.3        & 0.390     & 28.7    & 68.1    & 0.416    \\
SEA     \cite{pei2019semi}                     & 42.4        & 79.6    & 0.548     & 38.5        & 78.3        & 0.518     & 40.0    & 79.7    & 0.533    \\
\hline
GCN-Align   \cite{wang2018cross}             & 41.3        & 74.4    & 0.549     & 39.9        & 74.5        & 0.546     & 37.3    & 74.5    & 0.532    \\
KECG      \cite{li2019semi}                & 47.8        & 83.5     & 0.598     & 49.0        & 84.4        & 0.610     & 48.6    & 85.1    & 0.610    \\
MuGNN   \cite{cao2019multi}                & 49.4        & 84.4     & 0.611     & 50.1        & 85.7        & 0.621     & 49.5    & 87.0    & 0.621    \\
NAEA      \cite{zhu2019neighborhood}                & 65.0        & 86.7     & 0.720     & 64.1        & 87.3        & 0.718     & 67.3    & 89.4    & 0.752    \\
AliNet     \cite{sun2020knowledge}                 & 53.9        & 82.6     & 0.628     & 54.9        & 83.1        & 0.645     & 55.2    & 85.2    & 0.657    \\
\hline
GMNN    \cite{xu2019cross}               & 67.9        & 78.5    & 0.694     & 74.0        & 87.2        & 0.789     & 89.4    & 95.2    & 0.913    \\
RDGCN    \cite{wu2019relation}               & 70.8        & 84.6    & 0.746     & 76.7        & 89.5        & 0.812     & 88.6    & 95.7    & 0.911    \\
HGCN      \cite{wu2019jointly}               & 72.0        & 85.7    & 0.768      & 76.6        & 89.7        & 0.813     & 89.2    & 96.1    & 0.917    \\
NMN       \cite{wu2020neighborhood}                & \underline{73.3}        & \underline{86.9}     & \underline{0.781}     & \underline{78.5}        & \underline{91.2}        & \underline{0.827}     & \underline{90.2}    & \underline{96.7}    & \underline{0.924}    \\
\hline  \hline
\bf{RNM}                & \bf84.0        & \bf91.9   & \bf0.870        & \bf87.2       & \bf94.4     & \bf0.899         & \bf93.8        & \bf98.1   & \bf0.954     \\
\hline
\end{tabular}
\caption{Performance of different entity alignment methods}
\label{ent_align}
\end{table*}

\begin{table*}[]
\centering \small
\begin{tabular}{c|ccc|ccc|ccc}
\hline
\multirow{2}{*}{Models}    & \multicolumn{3}{|c|}{ZH-EN} & \multicolumn{3}{|c|}{JA-EN}  & \multicolumn{3}{|c}{FR-EN} \\
\cline{2-10}
 & Hits@1    &  Hits@10    & MRR     & Hits@1    &  Hits@10    & MRR    & Hits@1    &  Hits@10    & MRR  \\
\hline
\bf{RNM}                & \bf84.0        & \bf91.9   & \bf0.870        & \bf87.2       & \bf94.4     & \bf0.899         & \bf93.8        & \bf98.1   & \bf0.954     \\
\hline
RNM (-AP)              & 81.8        & 91.6    & 0.856     & 85.7        & 94.4        & 0.890     & 93.0    & 98.0    & 0.945    \\
RNM (-IS)               & 81.6        & 91.1    & 0.852     & 84.6        & 93.7        & 0.881     & 92.5    & 97.7    & 0.945    \\
RNM (-RM)              & 78.5        & 90.6    & 0.830     & 83.3        & 93.6        & 0.871     & 91.3    & 97.1    & 0.935    \\
\hline
\end{tabular}
\caption{Ablation study of the proposed model} \label{ablation}
\end{table*}

\subsubsection{Evaluation Metrics and Baselines}

The same as in previous work \cite{sun2018bootstrapping, yang2020cotsae}, 
we adopt Hits@$k$ and mean
reciprocal rank (MRR) as the evaluation metrics. Hits@$k$ measures
the proportion of correctly aligned entities ranked in the top $k$
list. $k$ is set as 1 and 10 as in previous work. MRR is calculated
as the average of the reciprocal ranks of the results. Higher
Hits@$k$ or MRR indicates the better performance of the model.

For comparison, we choose several competitive entity alignment
methods as baselines and classify them into three categories: (1)
TransE-based models, including MTransE \cite{chen2017multilingual},
IPTransE \cite{hao2017Iterative}, BootEA
\cite{sun2018bootstrapping}, AKE \cite{lin2019guiding}, 
and SEA \cite{pei2019semi}; (2) GCN-based
models which only utilize structural information, including GCN-Align \cite{wang2018cross}, KECG
\cite{li2019semi}, MuGNN \cite{cao2019multi}, NAEA
\cite{zhu2019neighborhood}, and AliNet \cite{sun2020knowledge}; 
(3) GCN-based models which employ entity name initialization,  
including GMNN \cite{xu2019cross}, RDGCN \cite{wu2019relation}, HGCN
\cite{wu2019jointly},  and NMN \cite{wu2020neighborhood}. 
Note that GMNN and NMN are models with subgraph matching.

\subsection{Experimental Results}

\subsubsection{Entity Alignment}

Table \ref{ent_align} shows the performance of different methods on
the entity alignment task. The results of Hits@1 and Hits@10 are in percentage (\%).
Numbers in \textbf{bold} denote the best
results among all models and the underlined ones denote the second
best results. The experimental results show that RNM significantly
outperforms all baselines on the three datasets. And it can achieve
that all the values of Hits@1 higher than 80\%, those of Hits@10
higher than 90\%, and those of MRR higher than 0.85. It is worth
noting that Hits@1 directly reflects the accuracy of alignment.
Thus, the outstanding results in Hits@1 further confirms the
effectiveness of the proposed model RNM.

Specifically, among all TransE-based models, BootEA performs the
best because it adopts a bootstrap strategy to iteratively
expand the seed alignments. This indicates that the iterative strategy can
significantly improve the performance of entity alignment. 
And for GCN-based models that only consider the structural information,
NAEA outperforms the others probably because it considers both neighboring
nodes and relations when representing entities. This confirms
that the relation information is important for entity alignment.

%is probably because GCNs are able to better capture the information
%from neighbor entities.

Moreover, NMN performs the best among all baselines. The
improvements may come from its neighborhood matching module.
However, the proposed model RNM further outperforms NMN by 10.7\%, 8.7\%,
3.6\% in Hits@1. This confirms that neighborhood matching with
relations can effectively improve the performance of entity
alignment.

\begin{figure*}[htb]
  \centering
  \subfigure[ZH-EN]{
    \includegraphics[width=0.218\linewidth]{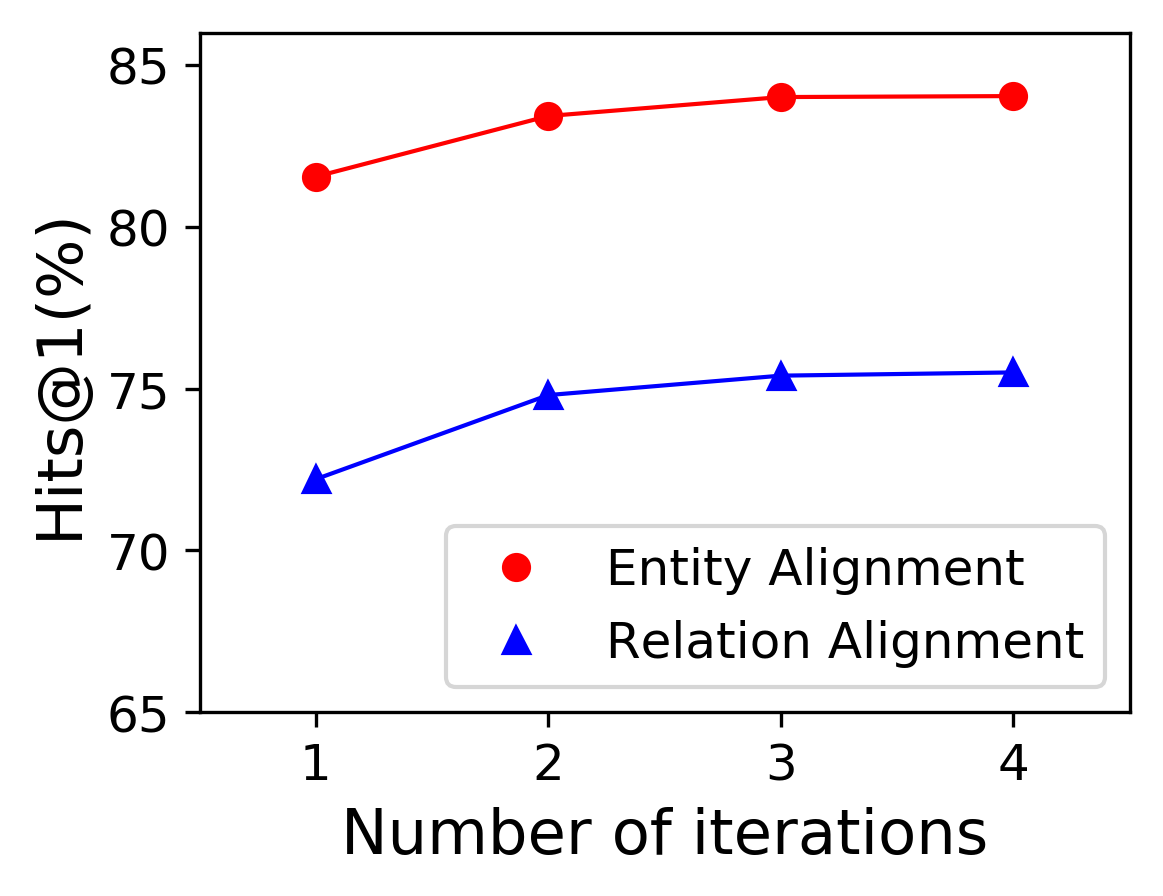}
  }
  \subfigure[JA-EN]{
    \includegraphics[width=0.218\linewidth]{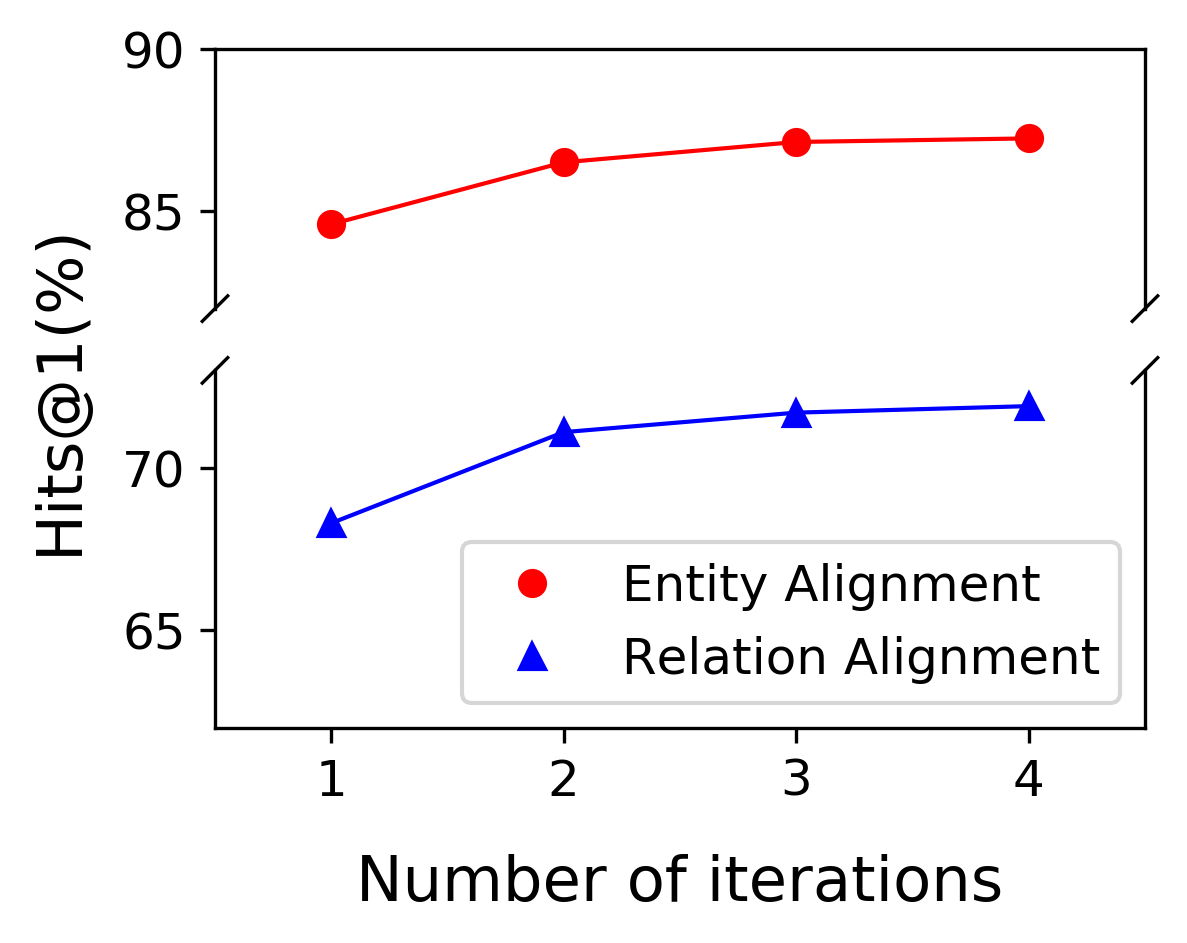}
  }
    \subfigure[FR-EN]{
    \includegraphics[width=0.218\linewidth]{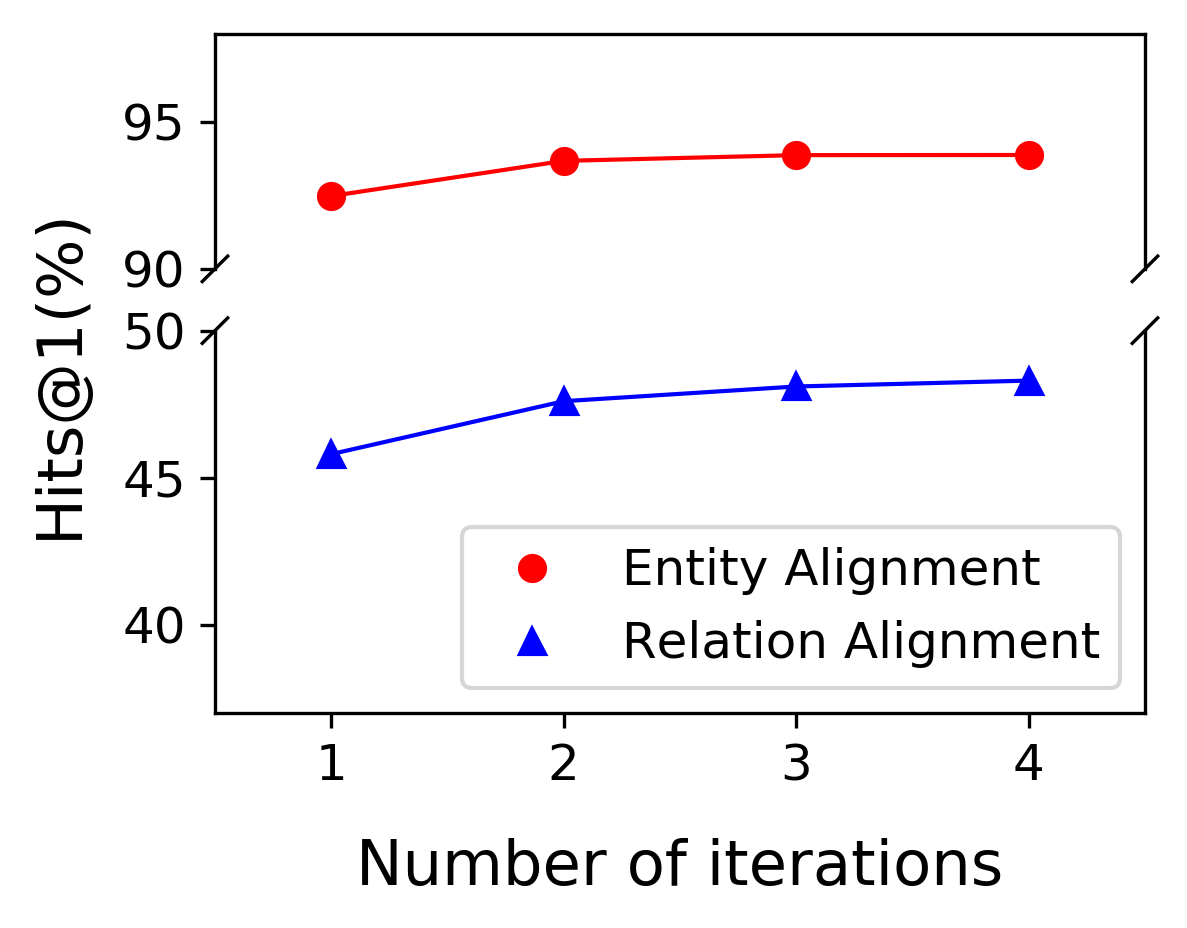}
  }
  \caption{Results of entity alignment and relation alignment w.r.t the number of iterations.}
  \label{iter}
  \vspace{-2mm}
\end{figure*}

\begin{figure*}[htb]
  \centering
  \subfigure[ZH-EN]{
    \includegraphics[width=0.218\linewidth]{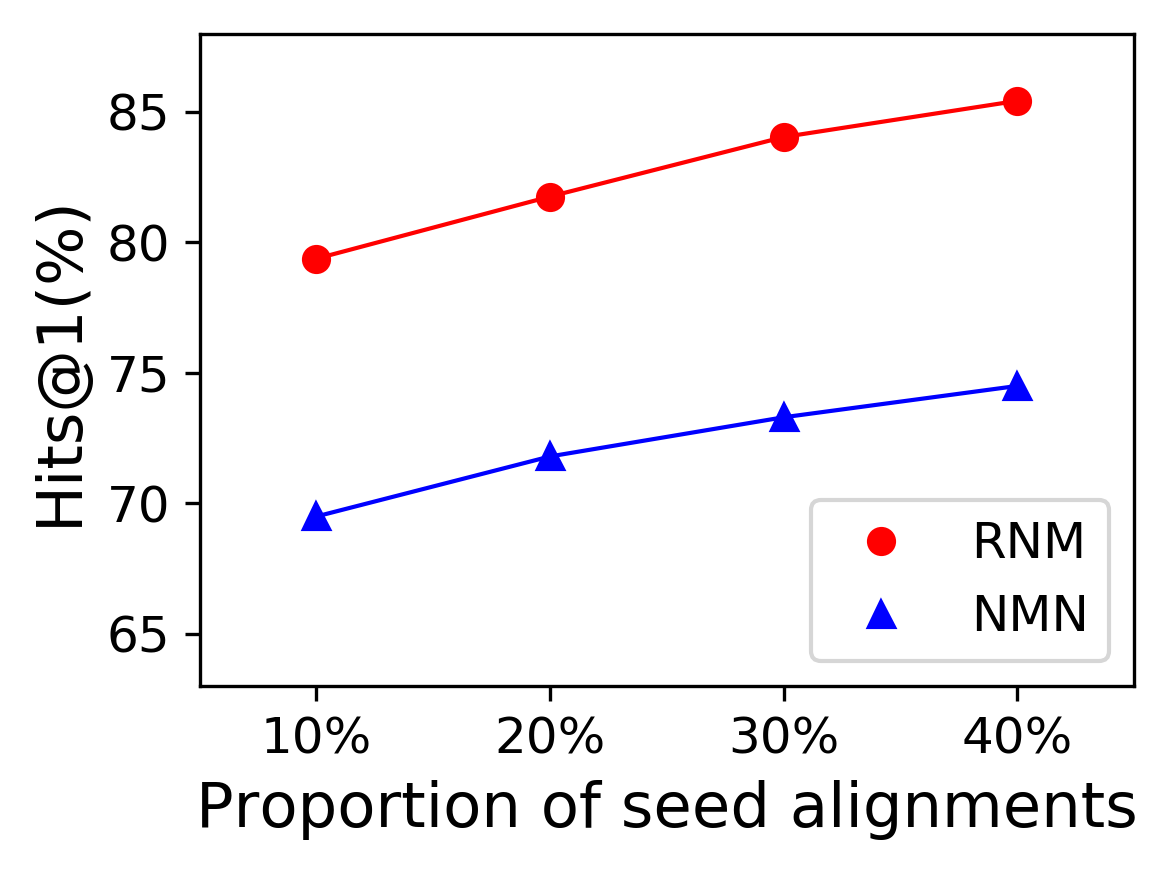}
  }
  \subfigure[JA-EN]{
    \includegraphics[width=0.218\linewidth]{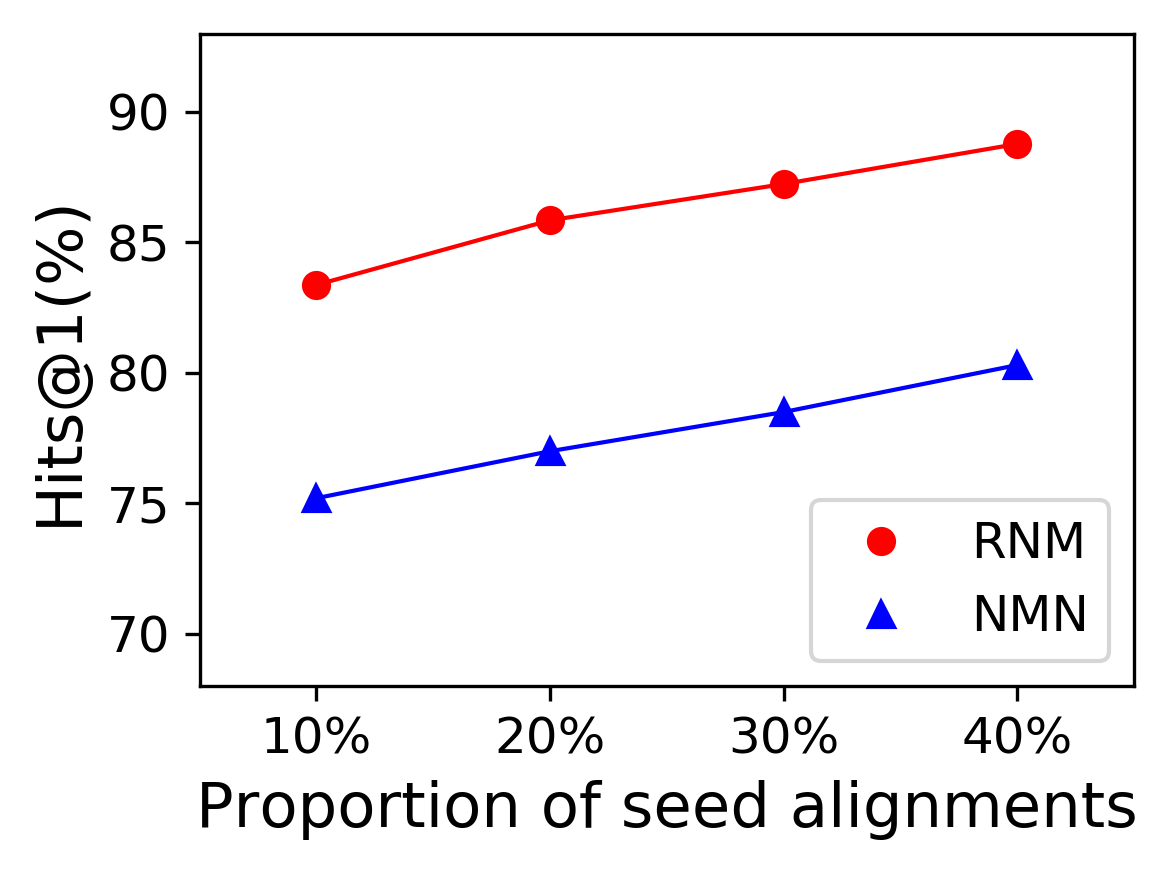}
  }
    \subfigure[FR-EN]{
    \includegraphics[width=0.218\linewidth]{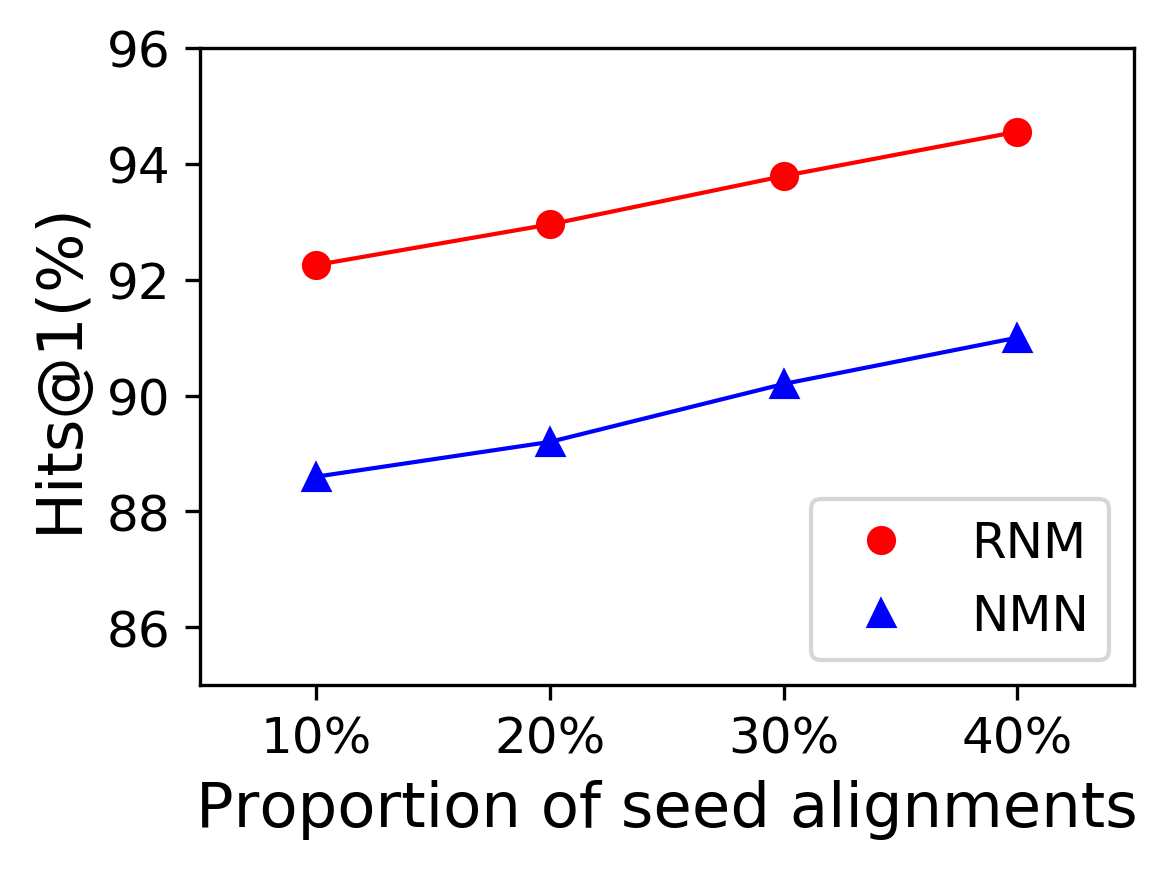}
  }
  \caption{Results of entity alignment w.r.t the proportion of seed alignments.}
  \label{seed}
\end{figure*}

\subsubsection{Ablation Study}

To evaluate the effectiveness of our designed modules, we construct
several ablation studies on the proposed model RNM and Table
\ref{ablation} shows the results. Specifically, (1) \textbf{RNM
(-AP)} denotes the model RNM without considering the alignment
probability (Eq. (\ref{ap})) in the relation-aware neighborhood matching
module; (2) \textbf{RNM (-IS)} denotes RNM without the
iterative strategy; (3) \textbf{RNM (-RM)}
denotes RNM without considering relations when matching neighborhood
for entity pairs. From the experimental results, we can observe that
the performance of RNM (-RM) drops the most, which confirms that it
is important to take into account the connected relations when
matching neighborhood for entity pairs. It is noted that RNM (-IS) consistently
outperforms all baselines in Table \ref{ent_align}, which confirms the effectiveness
of proposed relation-aware neighborhood matching module.
In addition, RNM (-AP) is 2.2\%, 1.5\% and 0.8\% lower than RNM in Hits@1 on three datasets,
which indicates that considering mapping properties of relations could further improve
the accuracy of entity alignment.

\subsubsection{Relation Alignment}

\begin{table}[h]
\centering \small  \resizebox{8.3cm}{!}{
\begin{tabular}{c|cc|cc|cc}
\hline
\multirow{2}{*}{Models}    & \multicolumn{2}{|c|}{ZH-EN} & \multicolumn{2}{|c|}{JA-EN}  & \multicolumn{2}{|c}{FR-EN} \\
\cline{2-7}
 & Hits@1    &  Hits@10    & Hits@1    & Hits@10   & Hits@1      & Hits@10  \\
\hline
MTransE-R                 & 3.0          & 8.9          & 2.7        & 10.2          & 3.3         & 14.6    \\
MTransE-PR              & 32.8        & 57.6        & 31.0        & 56.1        &18.9         & 44.3     \\
BootEA-R                   & 55.2        & 70.0        & 47.8        & 67.7       & 36.8         & 58.5     \\
BootEA-PR                & 45.3        & 85.4        & 41.4        & 79.8        & 30.2        & \underline{60.4}     \\
GCN-PR                    & 66.2        & 82.8        & 60.9        & 81.5        & 38.2        & 52.8     \\
GCN-JR                     & 70.2        & 82.8        & 63.9        & 81.8        & 42.0        & 53.8     \\
HGCN-PR                  & 69.3        & 84.5        & 63.1        & 81.3        & 41.5        & 54.3     \\
HGCN-JR                  & \underline{70.3}        & \underline{85.4}        & \underline{65.0}        & \underline{83.6}        & \underline{42.5}        & 56.6     \\
\hline
\bf{RNM}                & \bf80.6        & \bf87.1        & \bf74.5       & \bf84.6        & \bf49.5        & \bf62.5     \\
\hline
\end{tabular}}
\caption{Performance on relation alignment}
\label{rel_align}
\vspace{-1.5mm}
\end{table}

The proposed model RNM can not only be used for entity alignment but
also for relation alignment. Table \ref{rel_align} shows the
comparison results of different methods on the three datasets for
relation alignment. Note that the results of baselines are from the
reported data in \cite{wu2019jointly}, where -R denotes the original
model for relation alignment, -PR denotes the model that
approximates the relation representations using entity embeddings as
in \cite{wu2019jointly}, and -JR denotes the model that jointly
learns the embeddings of both entities and relations. From the
results we can observe that the proposed model RNM performs better
than all the baselines especially in Hits@1. Among the baselines,
BootEA achieves better results compared with MTransE due to its
bootstrapping strategy, while GCN further improves the performance
by incorporating the semantic information. The proposed model RNM
outperforms the best baseline model (HGCN-JR) by 10.3\%, 9.5\%, and
7.0\% in Hits@1 on the three datasets, respectively. The reason may
be that RNM aligns relations by matching the head and the tail
entities which can provide more evidence for relation alignment.
Moreover, these results confirm our assumption that the entity
alignment and the relation alignment can reinforce each other in our
model.

\subsubsection{Analysis}

Figure \ref{iter} shows the results of entity alignment and relation
alignment with different numbers of iterations for the proposed
model RNM. With the increase of iterations, the performance of RNM
on entity alignment and relation alignment raise accordingly. This
proves the effectiveness of the iterative framework of RNM and the
assumption that the entity alignment task and the relation alignment
task can reinforce each other for better performance.

%Moreover, the accuracy is nearly not improved from iteration 3 to
%iteration 4 probably because matching results seem to be stable and
%the distance between entities and between relations are no longer
%updated at this time.

Since RNM can implement the alignments in a semi-supervised manner,
we conduct several experiments on the entity alignment task with
different proportions of seed alignments, and results are shown in
Figure \ref{seed}. We choose NMN, which performs the best among baselines,  
as the comparison model. From the
results, we can observe that RNM outperforms NMN in all situations.
Even RNM with only 10\% seed alignments performs better than the NMN
with 40\% seed alignments on all the three datasets. This is because
RNM explores the useful information of relations when matching
neighborhood, and the iterative
strategy help to enhance the performance.

\section{Conclusion and Future Work}

In this paper, we propose a novel relation-aware neighborhood
matching model named RNM for entity alignment. In the model,
we jointly learn the embeddings of entities and relations.
Moreover, we propose to make use of the semantic information
and mapping properties of relations for better entity alignment.
In addition, we implement entity
alignment and relation alignment iteratively to reinforce each other 
in a semi-supervised manner. Finally, we evaluate
the proposed model on three cross-lingual KG datasets and empirical
results demonstrate the effectiveness of RNM.

In the future, we will study how to make use of the side information such as 
attributes \cite{zhang2019multi} and descriptions \cite{yang2019aligning}
to improve the accuracy of entity alignment.

\section{Acknowledgments}

This work was supported by the National Key Research and 
Development Project (Grant No.2020AAA0106600) and 
Peking University Education Big Data Project (Grant No.2020YBC10).

\bibliography{paper}

\end{document}